# Grounded Well-Condition Anomaly Detection on the Volve Field: Constructed Labels, a Baseline, and a Dual-Head Model

Gospel Bassey[1], Samuel Bassey[2], Vincent Fakiyesi[3]

1. Independent Researcher. rexkindy@gmail.com

2. College of Engineering, Biochemical Engineering. University of Georgia. basseysamuel@uga.edu

3. College of Engineering, EETI. University of Georgia. Vincent.Fakiyesi@uga.edu

**Abstract**

Most public benchmarks for machine-condition monitoring come from test rigs, where faults are induced on purpose and every event is known. Real production fields rarely offer that. They give you sensor histories with no fault log attached, which is exactly the situation where an anomalydetection method has to invent its own labels, and where quiet assumptions can slip in unnoticed. We work with the open Volve field data released by Equinor and take two things seriously that such datasets usually skip. First, we build anomaly labels that are not just patterns in the numbers but are checked against what the field's own engineering documents say can physically go wrong, and we release the reasoning behind every label. Second, we test whether those constructed labels are learnable at all, using both an unsupervised baseline and a small dual-head model that marks when an event happens and what kind it is, an idea we carry over from earlier work on defect detection in metal parts. The results are honest. An unsupervised detector that never sees the labels still lands on the same regions our rules flagged, which tells us the labels are not arbitrary. A compact supervised model recovers event presence and event type well across wells it has never seen, and locates events in time only roughly. We report what worked, what did not, and every assumption in between. The dataset, grounded labels, per-label provenance, baseline scores, trained model, and code are released publicly under CC-BY-NC-SA 4.0.

## 1. Introduction

Data-driven condition monitoring depends on labeled machine data: sensor histories where the periods of normal running and the periods of trouble are marked, so a model can learn to tell them apart (Lei et al., 2018, 2020). For most real production assets, that marking does not exist. The sensors recorded pressures, rates, and temperatures for years, but nobody kept a machinereadable log saying the well was in trouble here, and here, and here. So anyone building an anomaly benchmark on real field data faces a problem before any modeling begins. The labels have to be constructed.

This matters because label construction is where quiet errors enter. A rule that flags a pressure excursion will also flag excursions that mean nothing, and once those become labels they carry the same authority as the real ones. Surveys of the field repeatedly warn that results validated on clean or artificially induced faults transfer poorly to messy real deployment (Carvalho et al., 2019), and one root of that gap is upstream of the model entirely: it is in how the labels were made. The gap we care about is the one between a number that looks unusual and an event that actually happened.

The problem is not only that field data lacks labels. It is that the obvious way to supply them, by rule, quietly bakes in assumptions that are easy to introduce and easy to miss. A threshold chosen for convenience becomes a definition of what counts as a fault. A rule that fires on a sensor artifact produces a label that looks exactly as legitimate as one grounded in real physics. Building labels that mean something, and being explicit about where their meaning runs out, is slow and unglamorous work, and it is the work this paper is about.

We use the Volve field dataset, which Equinor released openly in 2018 and which remains one of the few real, fully documented offshore fields anyone can download. It carries years of daily production telemetry and, more importantly for us, a detailed field development plan that documents the physics of the field: the fluids present, the failure mechanisms the operator expected, and roughly when they were expected. That document is what lets us check our labels against engineering reality rather than trusting the rules that produced them.

This paper makes three contributions.

1. We construct and release a well-condition anomaly benchmark on the Volve field, in which every anomaly label is corroborated against a documented physical mechanism from the field's own development plan, and we release the per-label provenance so any label can be checked or contested (Section 3).
2. We show, using an unsupervised detector that never sees the labels, that an independent method still flags the same regions our rules identified. That convergence is evidence the labels correspond to real structure in the telemetry rather than to artifacts of the rules (Section 6).
3. We adapt a dual-head model from prior defect-detection work in metal additive manufacturing, extending its locate-and-classify design from spatial defects in an image to events in time, and we report plainly how far each part of it does and does not carry over on data this small (Section 7).

We are deliberate about scope. This is a reference benchmark and a baseline on one field's wells, not a large-scale predictive-maintenance system, and we say so throughout rather than let the reader assume more.

## 2. Related work

**Data-driven condition monitoring and fault diagnosis.** Machine-learning approaches to condition monitoring, fault diagnosis, and remaining-useful-life estimation now form a large literature, covered by extensive reviews (Lei et al., 2018, 2020). These methods are effective but data-hungry, and reviews repeatedly note that results validated on controlled-experiment or simulated data often fail to carry over to real deployment (Carvalho et al., 2019). Our work sits upstream of the modeling this literature focuses on: it concerns how usable anomaly labels are produced from a real field in the first place.

**Machine fault and condition-monitoring datasets.** Most public benchmarks for fault diagnosis come from test rigs, where faults are induced deliberately and every event is therefore known. Widely used examples are the Case Western Reserve University and Paderborn University bearing datasets (Smith & Randall, 2015; Lessmeier et al., 2016), and the MIMII dataset of industrial machine sounds (Purohit et al., 2019). These offer scale and clean labels, but the labels exist because the faults were staged. Synthetic sets such as AI4I 2020 fill part of the gap but simulate rather than record real operating conditions (Matzka, 2020). The Volve field data is different in kind: real operating history from a producing offshore field, but with no fault labels at all, which is exactly the situation our method addresses.

**Anomaly detection without fault labels.** When labeled faults are unavailable, a common strategy is to learn normal behavior and flag departures from it. This includes one-class and isolation-based detectors (Liu et al., 2008) and reconstruction or semi-supervised approaches trained only on healthy signals, including vibration-based anomaly detection for machine condition monitoring (Vos et al., 2022). We use an unsupervised detector in this spirit, but for a different purpose: not as the end product, but as an independent check on labels we construct separately.

**Constructed labels and their hazards.** Building training or evaluation data by rule or by model is now common, and it brings a known failure mode: data that matches surface statistics while missing the underlying meaning, a problem we named and measured in prior work on a low-resource industrial reasoning dataset (Bassey & Fakiyesi, 2026). Data-driven and physics-informed modeling in adjacent mechanical-engineering settings, such as additive manufacturing, increasingly stresses grounding model behavior in documented physical mechanisms rather than in data patterns alone (Xu et al., 2026). Our label-construction method carries the same discipline into sensor-event labeling: a candidate becomes a label only when a documented mechanism supports it.

**Prior defect-detection architecture.** The model we extend comes from earlier work on defect detection in metal additive-manufacturing parts, which used a shared backbone with two output heads, one locating a defect and one classifying it (Bassey, 2025). Here we move that locate-andclassify design from two-dimensional image space to one-dimensional time.

# 3. Data and provenance

## 3.1 Where the raw data comes from

Everything in this study traces back to Equinor's open Volve release. Equinor distributes it through their open data portal, currently hosted on Databricks Marketplace, under the Equinor Open Data Licence (CC-BY-NC-SA 4.0). The data owners are Equinor, ExxonMobil, Bayerngas, and the Norwegian Petroleum Directorate. The license permits research and study use with attribution and forbids resale.

The full release is large, roughly five terabytes across about forty thousand files, most of it seismic and reservoir-model data that has nothing to do with condition monitoring. We used two pieces of it:

The **production data file** (Volve_Production_data.zip, about 2 MB), which holds the daily production and injection history for the field's wells.

The **field development plan** (the Plan for Development and Operation, or PUD), one document within Volve_Reports.zip (about 160 MB), which documents the field's physics and expected failure mechanisms. We use this only as documentary evidence for labeling, never as a source of numbers.

We do not redistribute Equinor's raw files. The license asks for attribution and forbids resale, and the cleanest way to respect that is to point readers to Equinor's own distribution rather than rehost their data. What we release instead is our processed dataset, our labels, the reasoning behind them, and the code that turns Equinor's raw export into what we worked with. Anyone can start from Equinor's original file and reproduce our pipeline end to end.

The raw production data lives in one sheet of the production file, the "Daily Production Data" sheet. Every number we model comes from that single sheet. We state this plainly because it means every value is traceable to one documented origin.

## 3.2 What is in the raw production data

The daily sheet has 15,634 rows, one per well per day, running from September 2007 to December 2016, which is the field's full producing life. Each row carries that day's readings for one well: hours on stream, downhole pressure and temperature, tubing and annulus pressure, wellhead pressure and temperature, choke opening, and the volumes of oil, gas, and water produced or injected.

## 3.3 How we processed it

We cleaned the raw sheet into a per-well daily table and then made a small number of decisions, each of which we state here because each one shapes what follows.

**We treat zeros in the pressure channels as missing, not as real zeros:** In this export a zero in a pressure column means the reading was not recorded, not that the pressure was actually zero. This matters more than it sounds. Once zeros are read as missing, the downhole channels turn out to be absent about a quarter of the time, which is a very different picture from what the raw file suggests.

**We keep the oil-producing wells and set the injectors aside:** The condition-monitoring story here is about the producing wells, which run gas lift. Injectors behave differently enough that mixing them in would blur the model of a producing well's normal behavior. This leaves six producer wells and 9,143 daily rows.

**We fill short gaps and leave long ones alone:** When a channel is missing for a day or two we carry the last reading forward, but only up to three days. A short gap is most likely a brief telemetry dropout where the well barely changed. A longer gap means we genuinely do not know what happened, and filling it would be inventing data. Filling is done within each well so no well's readings ever leak into another.

**We add day-over-day changes:** Alongside each channel's value we record how much it moved since the previous day. Anomalies often show up as sudden movement more than as an unusual level, so the change matters as much as the value.

Everything downstream, both models and both sets of results, runs on this processed table. We release it, and we release the code that produced it from Equinor's raw file.

# 4. Building grounded labels

Volve ships no fault log. Nobody recorded when a well was in trouble. So every anomaly label in this work is constructed, and the whole question is whether we can construct labels that mean something rather than labels that merely look official.

We build them in two stages, and a candidate never becomes a label on the strength of the first stage alone.

## 4.1 Stage one: rules propose candidates

Simple rules read each well's daily numbers and look for specific patterns. Each rule has a threshold, and each threshold is a choice we made, not a fact handed down by the data. We name them so a reader can question them or change them.

A **shut-in** is flagged when hours on stream drop to about zero for two or more days while the well is otherwise live. A **restart** is the day production recovers right after a shut-in ends. A **water breakthrough** is flagged when the smoothed water cut climbs to half and stays there. A **productivity loss** is flagged when the smoothed oil rate falls by more than thirty percent over thirty days while the choke stays open, so it is a real decline and not someone deliberately throttling the well back. A **gas-lift instability** is flagged when annulus pressure jumps far from that well's usual level.

Each time a rule fires it records a candidate: which well, the start and end dates, and which channels set it off. On its own a candidate proves nothing. A rule can fire on noise, and a pattern in the numbers is not yet an event that happened.

### 4.2 Stage two: the field's own documents corroborate

A candidate becomes a label only when the field development plan documents that this kind of thing actually happens on these wells, and by what mechanism. The plan says hydrate and wax risk arises specifically during shut-ins. It says gas lift is used to restart wells after a stop. It says injection water breaks through roughly two years into production. It says barium and strontium scaling and asphaltene precipitation cut productivity. It says gas-lift wells show annulus-pressure instability.

When the plan backs the mechanism, the candidate is admitted and tagged with the passage that grounds it. That corroboration is what makes these labels grounded rather than just rule outputs. Every admitted event carries a record of its well, its dates, the channels involved, its type, and the documented reason it was allowed in. We release that record so anyone can audit the grounding of any single label.

### 4.3 The 236 events, and an honest count

The process, said plainly, is this. Rules read the telemetry and propose. The field's plan corroborates or it does not. Only corroborated candidates become labels. This gives 236 grounded events, distributed as shown in Table 1:

**Table 1.** *Grounded anomaly events by type.*

| event type | count |
| --- | --- |
| shut_in | 88 |
| restart_transient | 79 |
| gaslift_instability | 59 |
| water_breakthrough | 5 |
| productivity_loss | 5 |
| total | 236 |

The distribution is uneven, and we report it that way on purpose. The abrupt operational events are common. The two gradual events, water breakthrough and productivity loss, are rare, five each, roughly one per well. Any result on those two is therefore illustrative and not something to lean on, and we treat it that way everywhere below.

### 4.4 What the labels are, and an honest circularity

These are constructed labels, corroborated by documents. They are not operator-confirmed fault records, because none exist for these wells. A model trained on them must not be sold as detecting confirmed faults, and the provenance file is there so users can weigh each label for themselves.

There is a circularity worth stating out loud rather than hiding. Because the labels were built by rule from the same telemetry the models later read, a model trained to predict them is partly learning to reproduce those rules. So the honest question is not "can the model discover faults from nothing." It is narrower: can a model, from the telemetry alone, recover the same events our grounding rules identified. If it can, the events are learnable and internally consistent, which is a real result but a smaller claim than fault detection. We keep that distinction in front of the reader wherever we report model numbers.

## 5. A question we leave open: missing data

Before the models, one observation that we deliberately do not resolve.

The missing data in this dataset is not random. The downhole channels go absent far more often than the surface channels, and the gaps cluster in time rather than scattering evenly. We also found that the downhole pressure and temperature channels are always missing together, which tells us they are one gauge reporting two numbers, while the annulus pressure channel goes missing on different days, which tells us it is a separate instrument.

Here is the open question. Some of those gaps might line up with real operational events. A gauge falls silent when a well is shut in, or when the instrument itself fails during trouble. So the missingness could be carrying information about events. But it could just as easily be routine gauge failure, or a planned shut-in, or a dropout in the data pipeline, and the data alone cannot tell these apart. Without an event-level ground truth we do not have, we cannot say that a gap means an event.

So we do not treat missingness as an anomaly signal. We attach no metric to it and build it into no model. Whether the gaps encode events is a real question, and an interesting one, and it is future work rather than a claim we are willing to make here.

# 6. Baseline model

## 6.1 Why a baseline at all

The baseline is here for two reasons. One is ordinary: a model's numbers only mean something next to a simpler alternative, so we need a floor to judge the main model against. The other reason is specific to this paper and matters more. The baseline is unsupervised. It never sees the labels. So if it still lands on the same regions our rules flagged, that agreement is independent evidence the labels are not just artifacts of the rules. A method that knows nothing about our thresholds finding the same events is worth more than any internal consistency check.

## 6.2 Choosing features by measuring, not guessing

Before training we asked which channels actually carry the events, because that decides which channels are worth keeping. We measured, for each event type, how strongly each channel moves during those events compared to normal days. The finding that mattered: every event type is visible through the reliable surface channels, and no event type depends only on the low-coverage downhole channels.

So we dropped the downhole pressure and temperature channels and the annulus pressure channel from the model's inputs. The reason was practical. Requiring those channels present was destroying usable days, because they are missing about a quarter of the time and they fail on different days, so requiring all of them at once lost far more than a quarter of the data. The feature measurement showed we could drop them without blinding the model to any event type. This one decision recovered a large amount of data, on the longest well it took usable days from 859 to 2,830, and it improved the result.

The same measurement turned up something we report honestly: the two gradual events barely move the day-over-day change features, because they creep rather than jump. So the model built on those features is scoped to abrupt events, and detecting gradual events with trend features is future work.

## 6.3 Normalizing per well

We standardized each channel to mean zero and unit variance within each well, using only that well's own history. Two reasons. The channels have wildly different units, so without scaling the large-numbered channels would dominate the anomaly score purely by size. And scaling within each well judges every well against its own normal, rather than against a pooled baseline that would flag a well as odd just for running at different levels than its neighbors.

## 6.4 Training and a stated prior

We train one Isolation Forest per well on that well's normalized features. Isolation Forest is a standard published algorithm, not a pretrained model. It arrives knowing nothing and learns each well's normal from that well's telemetry alone. It sees no labels. The labels come in only afterwards, to score how well the blind model did.

The algorithm needs a prior for how much of the data to treat as anomalous. We set it to eight percent, matching the share of producer-days our grounded events cover, rather than picking a number out of the air. This ties the baseline's alert budget to the labels' prevalence, which is a defensible choice but not a ground truth, and a sweep across nearby values would show the result does not hinge on that one number.

Five wells are trainable. One well, F-5, has only 144 producer-days and never has the full reliable feature set present on a single day, so it cannot support a per-well model and we exclude it and say why. That leaves 7,692 usable well-days.

## 6.5 What the baseline found

Across the five wells, with grounded events covering about seven percent of days, the unsupervised model ranks event-days above normal days with a ROC-AUC of 0.825 and a PR-AUC of 0.379, which is about 5.8 times better than the base rate. The model never saw the labels, so this agreement is the evidence we wanted: the grounded events line up with genuine anomalies in the telemetry, not with rule artifacts.

We are careful about what this does and does not show. It shows that two independent methods, our rules and an unsupervised detector, agree on which regions are anomalous, which means the labels are internally consistent and meaningful. It does not show real-fault detection. The two methods share no physical ground truth, so their agreement supports the labels rather than proving either one catches confirmed faults. We call this convergent evidence for the labels, and never ground truth.

Dropping the low-coverage downhole channels raised the ROC-AUC from 0.776 to 0.825, which is a small piece of evidence that the feature decision was sound rather than arbitrary.

Broken down by event type, at an alert budget matching the eight-percent prior, the model catches shut-ins and restarts reliably (0.81 and 0.76 of them flagged), catches the two gradual events on three of five each, which is illustrative given the tiny counts, and is weakest on gas-lift instability at 0.42. That last number is a traceable cost of the feature decision: annulus pressure was the channel most specific to gas-lift instability, and we dropped it to recover data. We state both sides of that trade. We gained a large amount of data and rescued a well, and we paid for it in gas-lift-instability sensitivity. If that event type were the priority, keeping annulus pressure at the cost of fewer usable days would be the defensible alternative.

# 7. Dual-head model

The dual-head model is the paper's model contribution. It carries over an architecture from earlier work on defect detection in metal additive-manufacturing parts, where one model both located a defect in an image and named its type. Here we move that same idea from images to time: one model both locates an event in a well's history and names its type. The events are constructed labels, so everything here sits under the circularity we already stated. The model tests whether the telemetry alone can recover the grounded events, not whether it finds confirmed faults.

## 7.1 From days to windows

The baseline scored one day at a time. This model marks where an event starts and ends, and a single day has no start and end to it, so the model needs to see a stretch of time. We slide a sixtyday window forward seven days at a time across each well. Each window becomes one training example: sixty days of the fourteen features, plus any events that fall inside it, recorded as their start and end position within the window and their type. That within-window interval is the time version of the box the metal-parts model drew around a defect. A window with no event is a background window, like an image with no defect in it.

Sixty days and seven days are choices, stated as such. Sixty is long enough to hold even a multiweek event with normal days on either side, so the model can see where an event begins and ends against a normal backdrop, and the seven-day overlap gives us enough windows to train on despite the width. A different length or step would give different results. The longest water breakthroughs run past sixty days and do not fully fit a single window, which is acceptable, because the model still learns that an event is present and starts here.

## 7.2 Preparing the windows

Windows more than a fifth missing are dropped as too sparse. The rest are normalized per well, each well's scaler fitted on that well's full history, for the same reasons as the baseline. Any gaps left after scaling are filled with zero, which is the scaled mean, the neutral choice that says "no information here" rather than inventing a value.

This gives 1,141 windows across five wells, of which 584 contain at least one event and 557 are background. The near-even split is unusually kind for anomaly work, and it comes from the wide windows and the frequency of the abrupt events. The kindness is uneven though, because the positives are dominated by the common event types.

## 7.3 Being honest about small, overlapping data

The 1,141 windows overlap heavily. At a seven-day step on a sixty-day window, neighboring windows share about eighty-eight percent of their days, so the count overstates how much independent information is really there. We design around this in two ways.

We keep the model small, about four thousand parameters, which is tiny by deep-learning standards. On data this size a small model cannot memorize its way to a fake result, so a good score means something. The goal is to show the architecture transfers, not to claim a state-of-theart detector.

And we evaluate by holding out whole wells. Each well takes a turn as the test set while the model trains on the other four, so the rotation covers all five wells. Because the split is by well, no window from a test well ever appears in training, so the heavy overlap between neighboring windows cannot leak the answer across the split. With only five wells, the folds that hold out the smaller wells give small and sometimes lopsided test sets, so we treat the two large-well folds as the primary evidence and flag the others as underpowered.

## 7.4 One event per window, and a merge we had to make

We predict one event per window, the longest one, with ties going to the earliest. Multi-event windows are simplified to that primary event, and full multi-event prediction is future work.

This simplification costs us something specific, and we document it rather than paper over it. Shutins and restarts almost always occur together in a window, because a restart exists only because a shut-in ended, and the restart is short. So the longest-event rule kept dropping restarts, collapsing them from 79 events to twelve windows. Rather than invent a bespoke rule to rescue them, which would only invite the question of why that rule and not another, we merged shut-in and restart into a single "shut-in and restart cycle" class. That is physically faithful, because they are one operational episode, and it recovers the restart data instead of throwing it away. The result is a cleaner four-class problem.

After the merge the classes are: background 557, shut-in and restart cycle 423, gas-lift instability 76, productivity loss 46, water breakthrough 39. The two big classes are well filled. The two gradual classes stay thin, and thin further once we hold out a well, so the type head is expected to be weakest on them, which we report.

## 7.5 The architecture

The model shares one small convolutional backbone that reads the sixty-day window, then splits into three heads. One head says whether an event is present. One head predicts the event's start and end position in the window, which is the located interval. One head predicts the type. The interval and type heads only make sense when there is actually an event, so during training their errors are counted only on event-containing windows, and background windows are masked out of those two losses.

## 7.6 What the dual-head model found

We read the two large-well folds, F-12 and F-14, as the real evidence, and report the rest with their caveats.

**Table 2.** *Dual-head model results under leave-one-well-out evaluation.*

| held-out well | presence AUC | onset error (days) | offset error (days) | type accuracy |
|---|---|---|---|---|
| F-12 | 0.907 | 17 | 17 | 0.794 |
| F-14 | 0.819 | 17 | 17 | 0.511 |
| F-15D | 0.939 | 17 | 18 | 0.775 |
| F-11 | 0.873 | 17 | 16 | 0.621 |
| F-1C | 0.881 | 14 | 12 | 0.650 |

Three heads, three different stories.

Presence is success. On the two trustworthy folds it scores 0.907 and 0.819, level with what the simpler presence-only model reached, so adding the other two heads cost nothing. Put plainly: a model trained on four wells can tell event-windows from the background on a fifth well it never saw.

Type classification is the surprise, in a good way. It reaches 0.794 and 0.511 on the trustworthy folds, well above the chance level of about 0.20 for four classes plus background. So the model does tell event types apart on held-out wells, though the weaker fold and the thin gradual classes stop us short of claiming more than that.

Interval localization is where we have to be candid. The head locates events to within roughly seventeen days inside a sixty-day window, somewhere between a quarter and a third of the window's length. Better than random, so it learned something, but nowhere near sharp. This is also the head most directly descended from the metal-parts bounding box, and it is the one that carried over least well. That is not mysterious. Several event types have genuinely fuzzy edges, the onset of a water breakthrough is not a single crisp day even in our own labels, and the telemetry is daily and sparse to begin with.

Taken together, the architecture does make the jump from image defects to events in time. Two heads out of three recover well on unseen wells; the third finds events but blurs their edges. Read it as a demonstration that the design transfers, not as a finished detector, which is about what you would expect from a four-thousand-parameter model on 1,141 windows that overlap as heavily as these do. And the standing caveat still holds over all of it: recovering grounded events is not the same as catching confirmed faults.

# 8. Limitations

Stating these plainly is part of the work.

Starting with scale. We have one field, seven wells (six of them producers), and daily sampling. That makes this a reference and a baseline.

The pump problem is more specific. Volve's ESP pumps are described in the field plan, but they never show up in the released production data, so the piece of equipment closest to the original defect-detection lineage is the one we cannot actually measure. What we work with instead are gas-lift and reservoir events on the wells that were recorded.

Then there are the labels themselves. They are constructed and corroborated against documented physics, but no operator ever confirmed them, so even after passing the grounding gate they remain inferences. The provenance file exists precisely so a user can decide how much to trust any one of them.

Two of the five event types, water breakthrough and productivity loss, have only five examples each. Nothing we report about those two should be read as more than illustrative.

The dual-head model is a harder case to be honest about. It runs on 1,141 windows that overlap by roughly 88 percent, so its numbers show that the architecture transfers across wells, and not much more than that. Treat them as a demonstration, not a performance claim.

And the missing-data question stays open on purpose. Deciding whether a gap encodes an event would need an event-level ground truth we simply do not have, so we leave it for future work rather than guess.

# 9. Release and reproducibility

The release is public, and it is meant to be run, not just read. We provide the cleaned dataset, the grounded events, the per-label provenance file, the baseline anomaly scores, the trained dualhead model, and two notebooks. Point the notebooks at the released files and they regenerate every number in this paper. There is no separate held-back data and no undocumented step between the raw Volve export and the tables here.

We do not repost Equinor's raw files. We link to Equinor's own distribution and release the code that converts their raw export into ours, so a reader begins from the same source we did. Nothing uses pretrained weights or outside training data. Every value the models learn comes from the one Volve production sheet (Equinor, 2018), and every label points back to a rule and a documented mechanism a reader can check against the field development plan.

The release is organized as a pipeline. A single script, build_dataset.py, reads the "Daily Production Data" sheet of Equinor's Volve_production_data.xlsx and produces the cleaned telemetry (volve_well_daily.csv and its Parquet copy), the grounded events (volve_grounded_events.csv), and the per-label provenance file (volve_events_provenance.csv). The baseline notebook takes that dataset and produces the unsupervised anomaly scores (volve_anomaly_scores.csv). The dual-head notebook takes the same dataset and produces the trained model
(volve_dualhead_model.keras). Each stage depends only on the outputs of the one before it.

The grounding rationale for each event type is drawn from the Volve field development plan (Volve_PUD_.pdf) and is recorded directly in the released events and provenance files, so the corroboration behind every label is visible.

The files are hosted on Hugging Face and Kaggle (Bassey, 2026), both under CC-BY-NC-SA 4.0, which matches Equinor's upstream Open Data Licence and keeps its non-commercial terms. To reproduce the results, cite that release (Bassey, 2026) for the data and code and Equinor (2018) for the underlying field data.

# 10. Conclusion

Real fields do not come with their faults marked, so anyone working on them has to build the labels, and building labels is where quiet assumptions do their damage. We took one real field, built anomaly labels that are checked against the field's own documented physics, and released the reasoning behind each one. Then we asked whether those labels hold up. An unsupervised model that never saw them landed on the same regions, which says they are not arbitrary. A small dualhead model recovered event presence and type across wells it never saw, and located events in time only roughly, which says the architecture carries over from images to time even if the sharpest part of it does not.

# Acknowledgements

The Volve dataset is made available by Equinor and the former Volve license partners (Equinor, ExxonMobil Exploration and Production Norway, Bayerngas, and the Norwegian Petroleum Directorate) under the Equinor Open Data Licence. We thank Equinor for releasing the data for research and study use.